\documentclass[letterpaper, 10 pt, conference]{ieeeconf}  %

\IEEEoverridecommandlockouts                              %

\usepackage{graphics} %
\usepackage{epsfig} %
\usepackage{mathptmx} %
\usepackage{amsmath} %
\usepackage{amssymb}  %

\usepackage[]{todonotes}
\reversemarginpar
\usepackage{siunitx}

\usepackage[capitalise]{cleveref}
\usepackage{booktabs}
\usepackage{multirow}
\usepackage{subcaption}

\newcommand{\gaitphase}{g_\text{\%} }

\newcommand{\GFilter}{G_\text{pseudoInt}(s)}
\newcommand{\Fsum}{F_\text{sum}}
\newcommand{\Twindow}{T_\text{window}}
\newcommand{\Tsample}{T_\text{sample}
}

\newcommand{\nwindow}{n_\text{window}}

\title{\LARGE \bf
	Continuous locomotion mode recognition and gait phase estimation based on a shank-mounted IMU with artificial neural networks
}

\author{Florian Weigand$^{1}$,  Andreas Höhl$^{1}$, Julian Zeiss$^{1}$, Ulrich Konigorski$^{1}$ and Martin Grimmer$^{2}$ %
	\thanks{*The authors gratefully thank the Deutsche Forschungsgemeinschaft (DFG) for funding this work by the project with the reference number KO 1876/15-1 and GR 4689/3-1.}%
	\thanks{$^{1}$Florian Weigand, Andreas Höhl, Julian Zeiss and Ulrich Konigorski are with the Control Systems and Mechatronics Laboratory, Technical University of Darmstadt, Germany, {\tt\small fweigand@iat.tu-darmstadt.de}}%
	\thanks{$^{2}$Martin Grimmer is is with the Locomotion Laboratory, Institute of Sport Science, Technical University of Darmstadt, 
		Darmstadt, Germany,
	}%
}

\begin{document}

	\maketitle
	\thispagestyle{empty}
	\pagestyle{empty}

	\begin{abstract}
		To improve the control of wearable robotics for gait assistance, we present an approach for continuous locomotion mode recognition as well as gait phase and stair slope estimation based on artificial neural networks that include time history information. The input features consist exclusively of processed variables that can be measured with a single shank-mounted inertial measurement unit. We introduce a wearable device to acquire real-world environment test data to demonstrate the performance and the robustness of the approach. Mean absolute error (gait phase, stair slope) and accuracy (locomotion mode) were determined for steady level walking and steady stair ambulation. Robustness was assessed using test data from different sensor hardware, sensor fixations,  ambulation environments and subjects.
		The mean absolute error from the steady gait test data for the gait phase was \SIrange[range-units = single, range-phrase = --]{2.0}{3.5}{\percent} for gait phase estimation and \SIrange[range-units = single, range-phrase = --]{3.3}{3.8}{\degree} for stair slope estimation. The accuracy of classifying the correct locomotion mode on the test data with the utilization of time history information was in between \SI{98.51}{\percent} and \SI{99.67}{\percent}.  
		Results show high performance and robustness for continuously predicting gait phase, stair slope and locomotion mode during steady gait. As hypothesized, time history information improves the locomotion mode recognition. However, while the gait phase estimation performed well for untrained transitions between locomotion modes, our qualitative analysis revealed that it may be beneficial to include transition data into the training of the neural network to improve the prediction of the slope and the locomotion mode.
		Our results suggest that artificial neural networks could be used for high level control of wearable lower limb robotics.
	\end{abstract}

	\section{INTRODUCTION}
	
	Human locomotion is a complex process consisting of different locomotion modes, such as level walking or stair climbing, to navigate different environments. Each mode has specific support characteristics throughout the stride where the progress within a stride is called the gait phase. A specific event, e.g. the heel strike, is chosen as the beginning and the end of a stride. 
	The beginning of the stride corresponds to a gait phase of  \SI{0}{\%} and  the end to \SI{100}{\%} where the recurrence of the same event (heel strike) also defines the beginning of the following stride. 
	Both locomotion mode and gait phase contain useful information that can be used in the control of locomotion tasks. For wearable robotics, like powered prostheses or exoskeletons, knowledge of the users' locomotion mode and gait phase can be appropriate for optimal assistive support \cite{grimmer2014mimicking}.
	
	To obtain locomotion mode and gait phase, direct and indirect sensory information can be used. Direct sensory information requires some form of direct connection to the user, e.g. measuring muscle activity on the skin with surface electromyography (sEMG) \cite{Morbidoni2019}, or with implants, such as either an agonist-antagonist myoneural interface \cite{Clites2018AMI} or load cells to measure the interaction force between the device and the user \cite{Huang2011}. These approaches have limited user applicability \cite{Srinivasan2021AMI}, ease of use, and robustness \cite{Labarriere2020}, though they are still being actively researched.
	
	Indirect sensory information utilizes kinematic and kinetic measurements to estimate the locomotion mode and the gait phase. Inertial measurement units (IMUs) can be used to measure kinematic signals such as translational accelerations and angular velocities. Compared to sEMG, IMUs are robust and used in a wide range of everyday devices and can be easily attached to the user.
	Acquiring human kinetics such as joint torques requires measuring ground reaction forces (GRFs) in a laboratory setup. For everyday applications, measurement of GRFs is largely only available for wearable robotics and not for prostheses.

	Recently, research has focused on machine learning-based approaches for locomotion mode recognition \cite{Labarriere2020}. In \cite{intent_recognition_lower_limb_prosthesis_history_information}, a dynamic Bayesian network is proposed that uses distinct classifiers for locomotion mode recognition at certain predefined stride events. For each event a \SI{300}{\milli \second} window of data from 13 mechanical prosthesis-based sensors is used to calculate a likelihood for the current locomotion mode. To incorporate past predictions a prior probability for each gait mode is calculated based on the prediction of the last event. In~\cite{Spanias_2018} a \SI{300}{\milli \second} window at the beginning of each stride is used for the locomotion mode recognition based on sEMG data with dynamic Bayesian networks. A time window approach with a shank-based IMU was used in~\cite{bpnn_locomotion_mode_single_imu} where a decision tree based on neural networks as judgment nodes achieved high accuracy on classifying nine locomotion modes. However, transitions between the modes were not considered and the test data was not obtained in a different setup compared to the training data. Other approaches such as that in \cite{Huang2016} use multiple sensor systems, which are more challenging to incorporate in real-world applications. 
	
	Gait phase estimation can be realized by phase plane approaches \cite{Holgate2009, Quintero2017} and machine learning regression \cite{Seo2019, Kang2020, Weigand2020IFAC}.
	Seo et al. \cite{Seo2019} used recurrent neural networks (RNNs) with long short-term memory nodes (LSTMs) in combination with a shank-mounted IMU to estimate the gait phase for level walking. Results show that subject-specific training data improved the overall performance of the estimation. In \cite{Kang2020} artificial neural networks (ANNs) were used to estimate different walking speeds and the gait phase of level walking with a robotic hip exoskeleton that included two IMUs (thigh, trunk) and a hip encoder.
	
	In \cite{Weigand2020IFAC,Weigand2020Automed} we presented a proof of concept for a continuous gait phase estimation for level walking (LW), stair ascent (SA) and stair descent (SD) using ANNs with fully connected layers relying only on kinematic data from the shank as features. These features can be obtained with a single IMU mounted to the shank. However, we discovered that the performance of this approach was partially limited due to similar measurement data in the mid stance and the early swing phase.
	One way to overcome this limitation is to utilize past measurements or time history information. In \cite{Weigand2021Automed}, the use of time history information improved the estimation performance of both the gait phase and stair slope by over \SI{30}{\%} using a single ANN. The test data had mean absolute errors (MAE) of \SI{<3}{\%} for gait phase estimation and \SI{<4}{\degree} for stair slope estimation, though stair slope estimation was found to be quite noisy.
	
	The present work extends our previous use of an ANN for gait phase and stair slope estimation based on a shank-mounted IMU \cite{Weigand2020IFAC, Weigand2020Automed, Weigand2021Automed}. Specifically, we develop an additional ANN for continuous locomotion mode recognition without predefined stride events for level walking, stair ascent and stair descent to realize a complete high level control for a powered transtibial prosthesis.
	
	We hypothesize that it is possible to distinguish the locomotion modes LW, SA and SD with an ANN continuously throughout a stride. Our approach differs from the work of others that classify stride segments \cite{intent_recognition_lower_limb_prosthesis_history_information} or whole strides \cite{Gao2020} and therefore these approaches do not result in a classification for each sample.
	Further, as already observed for the gait phase estimation \cite{Weigand2021Automed}, we hypothesize that using time history information will improve the performance of the locomotion mode recognition.
	
	A laboratory-based dataset from a prior work \cite{Weigand2020IFAC, grimmer2020lower} that includes subjects without mobility impairment performing level walking and stair ambulation was used to train the ANNs.
	As the dataset was acquired in a laboratory setup it may not appropriately represent real-world scenarios. Therefore, the trained ANNs could have higher prediction errors if robotic hardware uses different sensors and sensor fixations from the ones used to acquire the training data. To allow for transferability to real-world scenarios and for use of different hardware, a mobile hardware setup was developed and used to collect test data in a real-world environment. As the data is additionally obtained from different subjects than in \cite{grimmer2020lower}, it can be considered as a true test scenario that is used to evaluate the robustness of the prediction of gait phase, stair slope and locomotion mode. We hypothesize that the performance for the test data from the laboratory experiment is slightly higher than that for the real-world environment due to the sum of differences in the real-world experiment.
	
	The training data for the ANNs only contains steady gait conditions. However, transitions between level walking and stair ambulation are important for wearable lower limb robotic control to avoid falls and injury, for which humans with impaired gait are at increased risk \cite{grimmer2019mobility}. Therefore, we qualitatively analyzed the behavior of the trained ANNs with respect to the unknown transitions as well. 
	We hypothesize that the gait phase during transitions will be detected with similar performance as that during steady gait. In addition we think that the estimated stair slope and the classified locomotion modes will gradually change with respect to the transition between two steady locomotion modes.
	
	\section{METHODS} \label{sec:methods}
	In the following sections the machine learning approach for the gait phase and stair slope estimation ANN is presented first. Then, the differences of the ANN regarding the locomotion mode recognition are stated and the data preprocessing for the ANNs is introduced.

	\subsection{Gait Phase and Stair Slope Estimation} \label{subsec:Regression}%
	
	For the gait phase and stair slope estimation we use an adapted ANN layout based on prior works \cite{Weigand2020IFAC,Weigand2021Automed} with fully connected layers and a moving time window for the input features. In comparison to RNNs this results in the utilisation of time history information without the need for more complex ANN architectures. 
	
	As in \cite{Weigand2020IFAC} only IMU measurements from the sagittal plane are used, resulting in two accelerations and one angular velocity. In combination with the pseudo-integration (\ref{subsec:methods:pseudo-int}) for each of these measured quantities and the time window a total of $6 \cdot \nwindow$ features are used as inputs for the ANN.
	
	All data used with the ANNs has a sampling frequency of \SI{200}{Hz}, resulting in a sampling interval $T_\text{sample}$ of \SI{5}{\milli \second}. The time window length is $\Twindow=\nwindow\cdot\Tsample$. The ANN is evaluated at each sample point resulting in updated output values every $\Tsample$.
	
	In contrast to our prior work \cite{Weigand2021Automed}, a cascaded structure for the hidden layer was not used, thus resulting in uniformly large layers. Hyperparameters were selected after a hyperparameter optimization (HPO) over a search space (\cref{tab:hyperparamterSpace}) on an eight-core desktop computer with GPU utilization (Nvidia RTX 2080) in Neural Network Intelligence (NNI) \cite{NNI} with a Gaussian process (GP) based Bayesian optimization. Selected parameters are presented bold in \cref{tab:hyperparamterSpace}. The optimal time window length of \SI{300}{\milli \second} from HPO matches that independently selected time window length of other approaches \cite{Kang2020, Young2013}.
	
	Start values of the network weights were randomly initialized with a default uniform distribution. The activation function of the hidden layers was a ReLu function. Due to the regression-type nature of estimating a continuous value like gait phase and slope the output layer has a linear activation function. The loss function for the training was a mean squared error and early stopping based on the validation loss with a patience of four was used.
	
	The ANN were implemented using \texttt{Python 3.8} \cite{Rossum2009} and \texttt{Tensorflow 2.2}. For training the \texttt{Adam}-Algorithm \cite{Adam2014} is used. 
	
	\begin{table}
		\caption{Parameter space for HPO for the ANN for gait phase and stair slope estimation. Selected parameters appear in bold. Time Window Length $\Twindow=\nwindow\cdot\Tsample$.}
		\centering
		\begin{tabular}{l r|l}
			\toprule
			\textbf{Hyperparameter}	            &           & \textbf{Range}\\
			\midrule
			Dropout Rate		                &           & [\textbf{0}\,,\,0.5]			\\
			Batch Size		                    &           & 128, 256, 512, 1024, \textbf{2048}, 4096			\\
			Layer Size                          &           & 32, 64, \textbf{128}, 256, 512, 1024        \\
			Hidden Layer Number                 &           & 1, 2, \textbf{3}, 4, 5, 6, 7, 8, 9        \\
			\multirow{2}{*}{Time Window Length} &$\nwindow$   & 1\,-\,75 (\textbf{60}) samples \\
			&$\Twindow$   & \SI{5}{\milli\second}\,-\,\SI{375}{\milli\second} ($\mathbf{300}$\,ms)\\
			\bottomrule
		\end{tabular}
		\label{tab:hyperparamterSpace}
	\end{table}
	
	\subsection{Locomotion Mode Recognition}
	The classification ANN used for locomotion mode recognition uses the same input space as in the ANN for the gait phase and stair slope estimation in \cref{subsec:Regression}. The output classes are LW, SA and SD as locomotion modes of interest.
	Note that we use a single classifier that is independent of the gait phase and thereby provides a prediction for each sample utilizing a sliding window of the most recent measurements (time history information). This is in contrast to \cite{sensors_activity_recognition_review} where it is stated that continuous classification requires multiple classifiers for different segments of a stride, as our approach eliminates the need for multiple classifiers. The hyperparameters for the classifier ANN in \cref{tab:hyperparamterSpaceRecognition} were selected by a grid search optimization based on the same parameter space as given in \cref{tab:hyperparamterSpace}. Only the time window length was selected based on the HPO of the gait phase and slope ANN to achieve consistency with the input feature dimension.
	\begin{table}
		\caption{Parameter space for HPO forthe ANN for locomotion mode recognition. Selected parameters appear in bold.}
		\centering
		\begin{tabular}{l|l}
			\toprule
			\textbf{Hyperparameter}	& \textbf{Range}\\
			\midrule
			Dropout Rate		    & 0-\,0.5 (\textbf{0.15})			\\
			Batch Size		        & 128, \textbf{256}, 512, 1024, 2048, 4096			\\
			Layer Size              & 32, 64, \textbf{128}, 256, 512, 1024      \\
			Hidden Layer Number     & 1, \textbf{2}, 3, 4, 5, 6, 7, 8, 9         \\
			Time Window Length $\nwindow$      & 60 samples \\
			\bottomrule
		\end{tabular}
		\label{tab:hyperparamterSpaceRecognition}
	\end{table}
	A hyberbolic tangent activation function was used for the hidden layers. Due to the classification-type nature of the locomotion mode recognition the output layer used a sotfmax activation function. The loss function for training was the categorical cross entropy and early stopping was used based on the validation loss with a patience of ten. The ANN for recognition was implemented in the same way as mentioned above for ANN for estimation.
	
	To determine the performance increase due to using the time history information, an additional ANN with no time window ($\nwindow=1$) was trained. 
	
	\subsection{Data Preprocessing}
	The following steps were used to preprocess the data. Note that these procedures are independent of the source of the dataset whether it be laboratory-based or real-world data.
	
	\subsubsection{Low-pass Filter}
	
	A second order low-pass Butterworth filter with a $f_\text{cutoff}=\SI{12}{Hz}$ was used to remove high frequency noise in IMU accelerations and angular velocities, which appear due to touchdown-related impacts as well as general measurement noise. To determine the cutoff frequency, the amplitude spectrum of the laboratory dataset was analyzed with a fast Fourier transformation (FFT). 
	
	\subsubsection{Pseudo-Integration} \label{subsec:methods:pseudo-int}
	The IMUs measure translational accelerations and angular velocities in three dimensions. Previously, it was shown that angle data can be beneficial for gait phase estimation \cite{Holgate2009}. Angle data can be calculated from IMU measurements through a Kalman-filter \cite{Rehbindera2001} or a complementary filter \cite{Gui2015}. 
	To realize the potential benefits of using angle information without the need to implement complex filters we introduced pseudo values of the angle, which are calculated by a pseudo-integration that consists of an integration followed by a first order high-pass \cite{Weigand2020IFAC}. This combination results in a drift-free qualitative representation of the integration. A similar approach is used in \cite{Holgate2009}.
	The combination of an integration followed by a first order high-pass filter can be condensed to a first order low-pass filter
	\begin{align*}
		\GFilter = \frac{1}{s}\frac{s}{Ts+1}=\frac{1}{Ts+1} 
	\end{align*}
	leaving the time constant $T$ of the filter to be set.
	The value of $T$ is set individually for angular velocities and translational accelerations by comparing the angle and translational velocities measured in the laboratory dataset to their pseudo counterparts. A compromise between a very strong (large $T$, less signal information) and a very weak (small $T$, very similar to the original signal, i.e. the angular velocity and the translational acceleration) low-pass filter must be identified. 
	The time constants for the pseudo angle and the pseudo velocity were set to $T_\text{angle}=\SI{1}{s}$ and $T_\text{velo}=\frac{1}{3}\si{\second}$, respectively.
	
\begin{figure}
		\centering
		\includegraphics[width=0.45\textwidth]{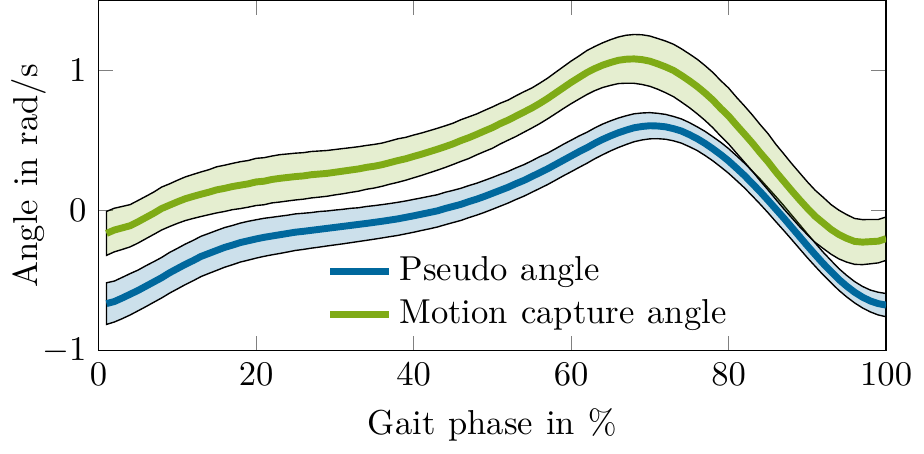}
		\caption{Comparison of pseudo and motion capture measured sagittal shank angle for steady LW from the entire laboratory dataset, including one standard deviation.}
		\label{fig:pseudo_angle_LW}
	\end{figure}

	As an example, \cref{fig:pseudo_angle_LW} shows the pseudo angle derived by pseudo-integration and the real angle measured in the lab. The major differences are the time lag of the pseudo value and the offset. The latter will be removed later in the process by the normalization of the input features. Therefore, the performance of the ANN should not be affected by the quantitative error of the pseudo values relative to real integrated values.

	\subsubsection{Gait Phase Separation}
	
	To improve the robustness and the usability for later use in a prosthetic high level control, the gait phase estimation is time-normalized within two sections of the stride. Touchdown and liftoff events are used to separate the stride into stance and swing phases. 
	We set a fixed gait phase of $\gaitphase=$\SI{63}{\percent} for the liftoff event as this is the middle of the liftoff gait phase range observed in \cite{grimmer2020lower}. The gait phase is then defined to linearly increase from \SI{0}{\percent} to \SI{63}{\percent} during the stance phase and from \SI{63}{\percent} to \SI{100}{\percent} during the swing phase. As a result the gait phase values depend on three events (touchdown, liftoff, next touchdown), which results in two linear sections with different slopes.
	This reduces differences occurring from variability between strides, individual timing or varying terrain. It is also beneficial for the ankle prosthesis application as stance and swing phase are often treated differently from a control perspective.

	\subsubsection{Gait Phase Transformation} \label{par:Gait_Phase_Transformation}
	In cyclic gait patterns, the gait phase percentage shows a discontinuity between the end of one stride ($\gaitphase=\SI{100}{\%}$) and the beginning of the following stride ($\gaitphase=\SI{0}{\%}$). To replicate the discontinuity within the gait phase estimator an infinite gradient would be necessary at the junction of two strides. It is plausible that the discontinuity in the gait phase results in an insufficient estimation performance around the beginning and the end of a stride. 
	To overcome this problem, we introduced a two-dimensional cartesian coordinate transformation \cite{Weigand2020IFAC}. This transformation was independently introduced in \cite{Seo2019} and \cite{Kang2020}, too.
	
	The gait phase value $\gaitphase$ can be considered an angle of a circle with a specific radius $r$, which is then transformed into cartesian coordinates
	\begin{align*}
		x &= r \cos\left(\frac{\gaitphase  2\pi}{100}\right), \\
		y &= r \sin\left(\frac{\gaitphase  2\pi}{100}\right)
	\end{align*}	
	to obtain a continuous representation of the gait phase with the cost of adding one extra dimension to the output values of the ANN for the gait phase. The estimated output values must be transformed back with a post-processing step to obtain the gait phase estimation as percentage of the stride.
	
	The two values $x$ and $y$ representing the gait phase and the stair slope are the three outputs of the estimation ANN.
	
	\subsubsection{Normalization}
	A normalization of input features and output labels improves the optimization stability during ANN training \cite{santurkar2018does}. 
	A naive normalization approach would be based on the mean and standard deviation (STD) of the entire training data. 
	During preliminary tests with data measured from sensor hardware other than that from which the training data were acquired, the ANNs were found to have poor estimation and recognition quality. 
	We explain this by deviations in the sensor orientations and mounting height on the shank from variable mounting of the IMUs during the real-world environment experiments. These deviations result in IMU measurements of differing amplitudes.
	In this case, we decided to normalize the input features of each real-world experiment dataset based on the mean and the STD of that specific experiment, which resulted in much better performance of the ANNs. The normalization values can easily be obtained for a new sensor system with a few measured strides. 
	Different locomotion modes exhibit different values for the mean and the STD. Therefore, using the mean and the STD of the complete dataset makes the normalization dependent on the distribution of the different locomotion modes. For this reason we only use steady LW data for normalization.
	
	\section{EXPERIMENTS}
	The study protocols for the laboratory and the real-world experiment were approved by the institutional review board of Technical University of Darmstadt, Germany. All subjects provided written informed consent in accordance with the Declaration of Helsinki. 
	
	\subsection{Laboratory Dataset}\label{subsec:Methods:DatasetStais2018} %
	The laboratory dataset used in this work was previously recorded during a stair walking experiment at the Locomotion Laboratory of the Technical University of Darmstadt \cite{grimmer2020lower}. 
	
	During the experiment twelve subjects (age: $\SI{25.4\pm4.5}{years}$, height: $\SI{180.1\pm4.6}{cm}$ and mass: \SI{74.6\pm7.9}{kg}, all male) were recorded while walking on an instrumented track that included a staircase and a level area before and after the staircase. Force plates (Kistler, Switzerland) were mounted within the track and staircase to capture GRFs.
	Each subject was equipped with combined sEMG and IMU sensors (Delsys Trigno, US), from which only the shank IMU data is of interest for this work. 
	
	\subsubsection{Experimental Protocol}
	Prior to recording, subjects could familiarize themselves with the instrumented track, and force plates were arranged individually to match the preferred step length. 
	Each subject walked along the track, ascending and descending the staircase ten times for each of three different stair slopes (\ang{19}, \ang{30} and \ang{40}).
	
	Following the experiment the vertical GRF and IMU data (translational accelerations and angular velocity) were extracted and the locomotion mode and the stair slope values were set according to the experimental protocol. The gait phase values had to be calculated based on the GRFs. The beginning and the end of a stride were timed by the touchdown event and the following touchdown of the same foot, respectively. Both touchdown events were determined by the rise of the vertical GRF in the corresponding force plate. The gait phase was set as a percentage of stride duration based on the beginning (\SI{0}{\%}) and end (\SI{100}{\%}) of the stride as well as the stance-swing phase division previously described.
	
	In total 2128 strides of LW, 1067 strides of SA and 1415 strides of SD were evaluated from the twelve subjects. All signals were equalized to a sample frequency of \SI{200}{Hz}.
	
	Due to the data being measured from individual subjects the data is split into training, validation and test data from nine, two and one of the twelve subjects, respectively. This process was based on the results of a cross-validation in \cite{Weigand2020Automed}. This split by subjects is also intended to provide a test of transferability to different individuals..

	\subsection{Mobile Hardware Setup and Real-World Dataset} \label{subsec:Methods:FRIMU} %
	
	\begin{figure*}
		\centering
		\includegraphics[width=0.65\textwidth]{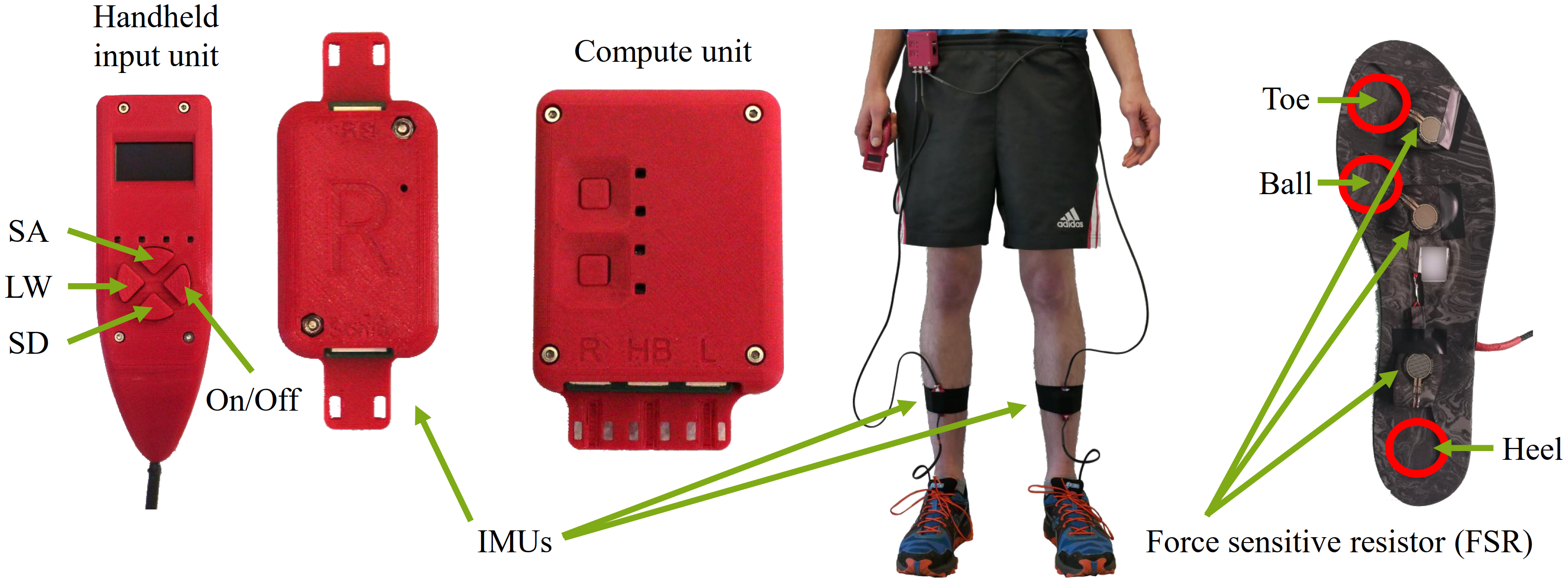}
		\caption{Overview of the mobile FRIMU2 device. From left to right: Handheld input device, IMU, hip-mounted compute unit and the underside of the insole with force sensitive resistors. The red circles indicate the mounting position of the FSRs, which are folded back to be visible in the picture.}
		\label{fig:FRIMU2_Complete}
	\end{figure*}

	The mobile hardware device FRIMU2 (\textbf{F}orce sensitive \textbf{R}esistor \textbf{IMU}, \cref{fig:FRIMU2_Complete}) enables us to perform experiments that are independent of the laboratory. It is used to acquire test data for the ANNs from different sensor hardware in a real-world terrain to evaluate the performance and robustness of the trained ANNs.

	The FRIMU2 contains an IMU-Sensor (Pololu MiniIMU9v5) at each shank for collecting kinematic measurements. Three Force Sensing Resistors (FSR, Interlink FSR 402 short) on each foot are used to measure ground contact of each foot to determine the beginning and the end of a stride. With the hand-held control unit the user can input the current locomotion mode by pressing the respective button (LW, SA, SD). With manually-measured stair slopes used during a data collection session, we have all of the necessary information to specify the outputs of the ANNs (gait phase, stair slope, locomotion mode). 
	
	\subsubsection{Ground Contact Determination}
	To obtain the correct gait phase, the FSR data need to be further processed. %
	First, the sum of all three FSRs per foot
	\begin{equation*}
		\Fsum =F_\text{heel}+F_\text{toe}+F_\text{ball}
	\end{equation*}
	is calculated, which qualitatively resembles the curve of the vertical GRF during ground contact but with a different scaling. As only the touchdown and liftoff are of interest to determine the gait phase, a qualitative representation of the vertical GRF is sufficient. 
	
	Secondly, similar to the laboratory dataset containing vertical GRF, the FSR data allows us to determine the touchdown and liftoff by the rise and fall of $\Fsum$. 
	A threshold value is set by manually inspecting $\Fsum$ for each foot and selecting the threshold based on the noise level during the swing phase. One threshold for each leg is specified for each individual experiment to account for subject-specific differences (e.g. tightness of the shoe or changes in FSR placement and subject weight).

	\subsubsection{Data Processing}
	
	The FRIMU2 contains an IMU at each shank that records translational accelerations and angular velocity. The IMU values are processed in the same manner as in the laboratory-based dataset.
	
	A cleaning procedure is used to remove artifacts such as irregular long or short strides and instances of turning on the spot or standing. A stride is removed if one of the following three criteria is met. First, the maximum sagittal angular velocity is below $|1|\,\si{\degree \per \second}$. Second, the maximum vertical translational acceleration is below $|0.1|\,\si{\meter \per \second}$. Third, the total stride length is smaller than \SI{0.8}{\second} or larger than \SI{1.4}{\second}.

	\subsubsection{Experimental Protocol}
	Test data of three subjects (age: $\SI{28.3\pm3.8}{years}$, height: $\SI{187\pm5}{cm}$ and mass: \SI{77.3\pm11,0}{kg}, all male) was collected with FRIMU2. 
	
	Each subject walked in a stairwell of a five-story building that included ten flights of stairs with a \SI{26}{\degree} stair slope. Subjects had the FRIMU2 device mounted as shown in \cref{fig:FRIMU2_Complete}. The locomotion mode was input by the subject during walking via the hand-held control unit by pressing the up, down or left button for SA, SD or LW, respectively. The subjects were told to press the button while the leading limb was in the swing phase when entering or exiting each staircase.
	
	Each experimental trial had a variable duration due to the subjects' preferred walking speed; trials lasted \SI{220}{\second}, \SI{230}{\second}, \SI{220}{\second} for subject 1, 2, and 3, respectively.
	Each subject was given time to become acquainted with the hand-held control unit and the staircase.
	The LW strides in the real-world experiment provide an additional challenge to the trained ANNs; while the ANNs were trained with data from a straight instrumented track in the laboratory experiment, the real-world staircase layout included a \SI{180}{\degree} turn with a radius of approximately \SI{3}{\meter} each half-flight of stairs. 
	This difference from straight locomotion creates a gait disturbance and therefore offers insights on the robustness of the ANNs.
	
	In addition, FRIMU2 data of one subject was measured while walking up and down a long, level hallway. This experiment allows us to investigate LW in more detail without the subject turning around as in the five-story building staircase.

	\section{RESULTS \& DISCUSSION}
	\begin{table*}
		\caption{Error measures for gait phase estimation, slope estimation and locomotion mode recognition with and without time history.}
		\centering
		\begin{tabular}{l c r|c c c | c c c c}
			& &            & \multicolumn{3}{c}{Laboratory dataset} & \multicolumn{4}{c}{Real-world dataset} \\
			&$\Twindow$  &  Measures          & Train  &   Val    & Test      &  Sub1   &  Sub2  &  Sub3    & Hallway Sub3\\
			\midrule
			Gait Phase	& \SI{300}{ms} & MAE in \%	                     &  1.2   & 1.6      & 2.0       & 2.5           & 2.6          & 2.2             & 3.5 	\\
			Stair Slope	& \SI{300}{ms}&MAE in \si{\degree}          &  2.2   & 3.0      & 3.3       & 2.6           & 3.4          & 3.8             & 1.0 	\\
			Locomotion Mode & \SI{300}{ms}  & Accuracy in \%    & 99.98  & 99.74    & 99.67     & 99.57         & 98.80        & 98.60           & 98.51  	\\
			Locomotion Mode & - &Accuracy in \% & 99.39& 94.41 & 97.91  & 96.39 & 95.14 & 94.90 & 97.82 \\
		\end{tabular}
		\label{tab:quantitative_Results_ANNs}
	\end{table*}
	
	The mean absolute error (MAE) of the gait phase and the stair slope estimation from the ANN were calculated as quantitative performance measure. The gait phase value is obtained by transformation from the cartesian output values of the ANN as described in \cref{sec:methods}. 
	The performance of locomotion mode recognition is quantified by accuracy of the predicted value relative to the true value.
	
	To analyze steady gait conditions and to compare real-world data with the laboratory-measured data, transition data were excluded from the real-world test dataset. Since humans need approximately one stride duration for the longest transition (SA to LW) \cite{grimmer2020lower}, we decided to exclude $\SI{1.29}{\second}\,(\SI{258}{samples})$ before and after the manually-entered locomotion mode change entered by the subjects. This period is twice as long as the longest stride duration observed in \cite{grimmer2020lower}.
	This post-processing step should leave only steady locomotion data if the locomotion mode change was input anytime during the transition phase. Due to the nature of the human-based input errors (remaining transitions within the steady gait data) can not be ruled out completely by this approach. For the laboratory dataset, from which only steady LW, SA and SD strides were included, no transition exclusion is necessary.

	For the laboratory dataset, gait phase and stair slope show a slight increase in MAE for the validation and test data relative to the training data \cref{tab:quantitative_Results_ANNs}. Locomotion mode accuracy declines slightly in the validation and test data relative to the training data.
	Such results could be expected as both datasets contain subject-specific gait unknown to the ANNs from training.
	
	Comparing the real-world test data to the laboratory test data shows slight increases for the MAE of the gait phase and slight decreases for the accuracy of the locomotion mode for all three subjects. Comparing the same data for stair slope only subject 1 has a decrease in MAE, while subjects 2 and 3 have slight increases. Such a result is in line with our hypotheses and demonstrates the robustness of the approach considering the differences in the laboratory and real-world experimental setup. These differences include sensor hardware, sensor placement and a real-world staircase with a difference in slope and turns between flights of stairs.
	
	Despite the differences between the real-world and laboratory setups, the MAEs for the gait phase estimation are still better than our previous results for the laboratory dataset in \cite{Weigand2020IFAC} where no time history information was utilized for the input features. This highlights the benefits of using the time history approach for gait phase estimation. 
	
	Locomotion mode recognition performance is close to perfect for the laboratory dataset with little differences between the training, validation and test data. The performance decrease with real-world test data is lower than the decrease in test data performance for the LSTM-based ANNs in \cite{Sherratt2021}. This result is promising as \cite{Sherratt2021} collected training and test data using the same hardware setup.
	
	The accuracy of the locomotion mode recognition for hallway data with subject 3 is similar to that accuracy achieved with combinations of stair ambulation and level walking. Therefore, we assume that stair ambulation has similar recognition quality as level walking and that level walking while turning on the landings between half-staircases has no negative impact on the results.  
	The MAE of the gait phase in the real-world hallway data was slightly higher than in the real-world stair ambulation data and in the laboratory data. We do not have an explanation on the reason for this finding but it is possible this could be attributed to differences in locomotion speed, which was not controlled in this trial.  
	
	As hypothesized, the use of time history information for the input features improves locomotion mode recognition for both datasets. Here, the additional information of the input features allows for a better distinction between conditions. As such information can be easily integrated without additional sensors, without more complex machine learning algorithms and without a major increase in computation cost, we highly recommend including time history information.

\begin{figure*}
		\centering
		\begin{subfigure}[b]{0.5\textwidth}
			\centering
			\includegraphics[]{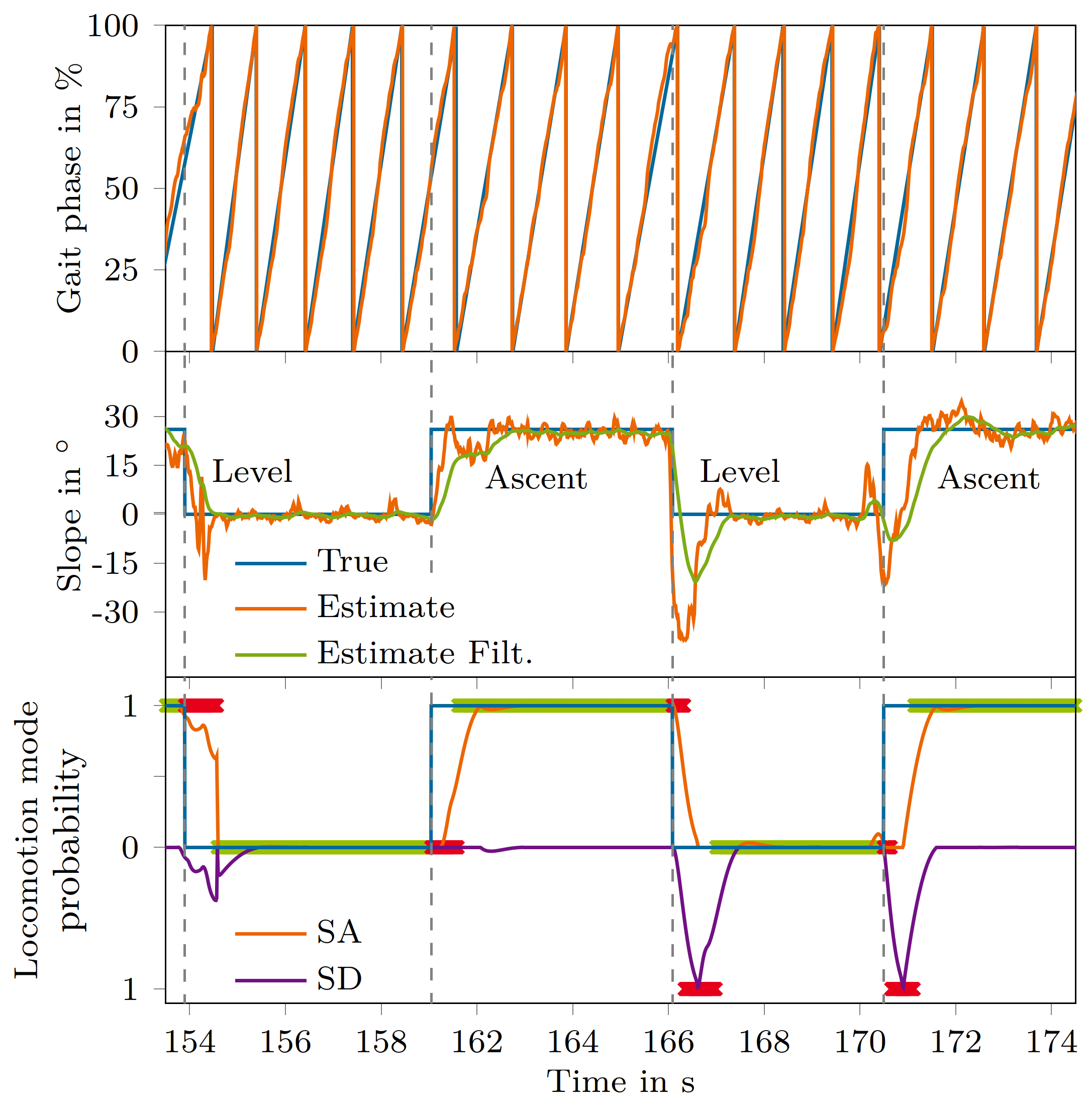}
			\caption{Subject 1 during ascent.}
			\label{fig:Complete_FRIMU2_JZ_rtm_Trial1_Ascent}
		\end{subfigure}
		\begin{subfigure}[b]{0.49\textwidth}
			\centering
			\includegraphics[]{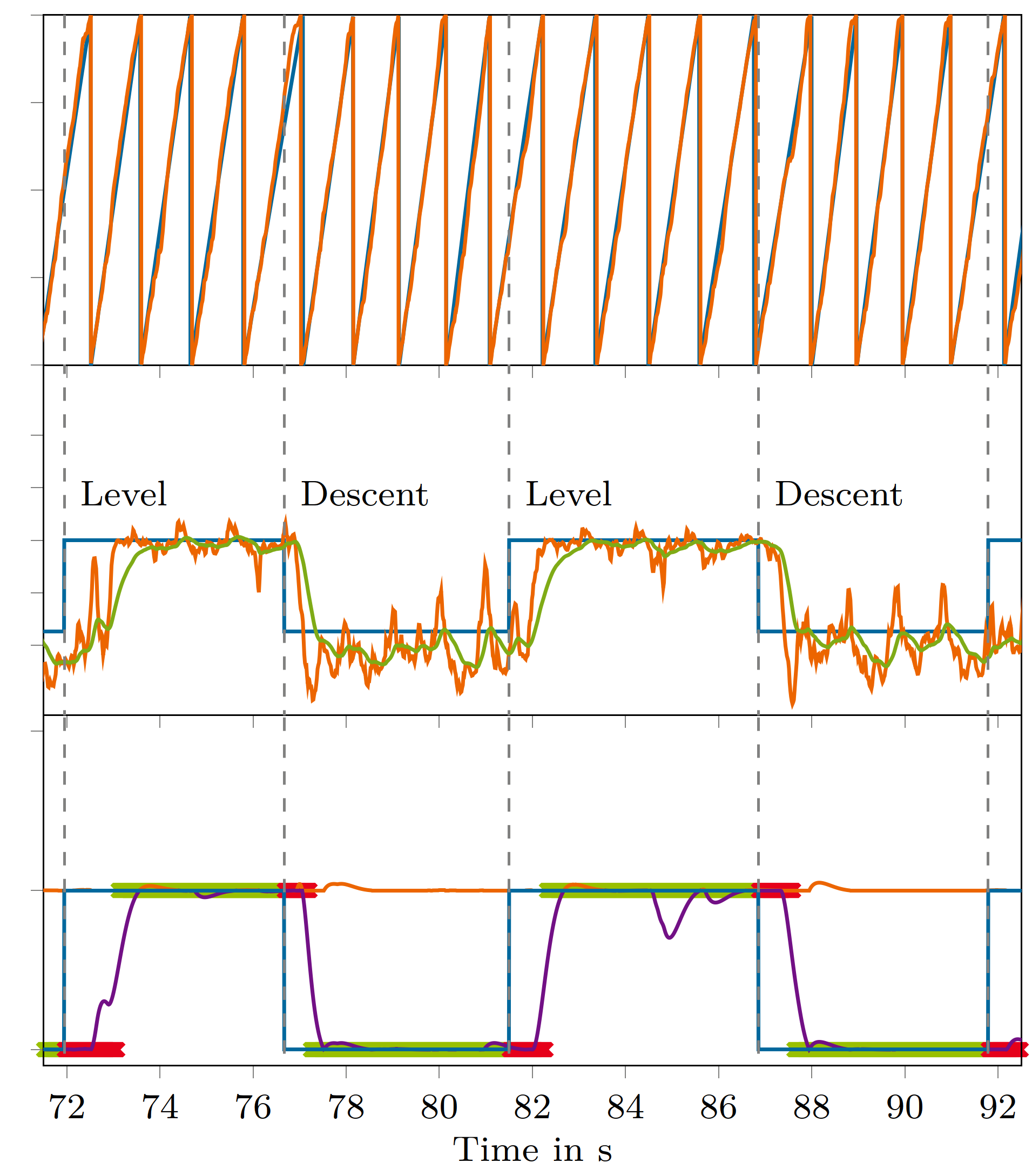}
			\caption{Subject 2 during descent.}
			\label{fig:Complete_FRIMU2_SK_rtm_Trial1_Descent}
		\end{subfigure}
		\caption{Example results of gait phase (top) and slope estimation (middle) and locomotion mode recognition (bottom) for test data measured with FRIMU2 in the five-story staircase. Stair slope estimation is presented as the original and filtered (for visualization purposes, 3rd-order, Butterworth low-pass, $f_\text{cutoff, slope}=\SI{0.5}{\hertz}$) estimation of the ANN. To improve visualization of the locomotion mode recognition the probabilities of each class are filtered as well (2nd-order, Butterworth low-pass, $f_\text{cutoff, class}=\SI{80}{\hertz}$). Green and red markings in the lower plot indicate correct and incorrect recognition of the locomotion class, respectively. The dashed lines indicate the subject's input timing of the locomotion mode change. The probability for LW is not shown for clarity, but it has similar trajectories. }
	\end{figure*}

	Since the transition timings in the real-world experiment with FRIMU2 are input by the user (dashed vertical lines in \cref{fig:Complete_FRIMU2_JZ_rtm_Trial1_Ascent} and \cref{fig:Complete_FRIMU2_SK_rtm_Trial1_Descent}), the accuracy of the timings is uncertain. A qualitative inspection of the results was therefore performed to evaluate the ANN performance with time history information for transitions that were not used in training. \cref{fig:Complete_FRIMU2_JZ_rtm_Trial1_Ascent} and \cref{fig:Complete_FRIMU2_SK_rtm_Trial1_Descent} are used to represent these qualitative results.

	\cref{fig:Complete_FRIMU2_JZ_rtm_Trial1_Ascent} and \cref{fig:Complete_FRIMU2_SK_rtm_Trial1_Descent} show \SI{20}{\second} from a \SI{220}{\second} and \SI{230}{\second}  real-world trial for the right leg of test subject 1 during stair ascent and the left leg from subject 2 for stair descent, respectively. \cref{fig:Complete_FRIMU2_JZ_rtm_Trial1_Ascent} and \cref{fig:Complete_FRIMU2_SK_rtm_Trial1_Descent} contain multiple transitions between locomotion modes and include the LW portions during the \SI{180}{\degree} turn in the half-story staircase. 
	
	The gait phase estimation matches our hypothesis with an almost similar quality for the steady strides and the transition strides. The estimated gait phase starts at \SI{0}{\percent} and reaches \SI{100}{\percent} without having unexpected behavior like sudden jumps in between. These results reflect the robustness of the trained ANN when facing test data with different sensor hardware, a different stair height, different sensor placement, transitions, and turns between stair segments. 
	
	Small consistent overestimations of the gait phase occurred during the transition from SA to LW at $t\approx\SI{166}{\second}$ (\cref{fig:Complete_FRIMU2_JZ_rtm_Trial1_Ascent}) and for LW to SD between $t=\SI{76}{\second}$ and $t=\SI{78}{\second}$ (\cref{fig:Complete_FRIMU2_SK_rtm_Trial1_Descent}). For the LW to SA and the SD to LW transitions no visible deviations occurred.
	
	In contrast to our hypothesis, the stair slope estimation revealed inconsistent and false estimations for the LW to SA and the SA to LW transitions, while the LW to SD and SD to LW transition show estimations close to the expected results of having a transition within the estimated slope. For stair ascent, for about one second the ANN falsely estimated a negative slope as can be seen at $t=\SI{166}{\second}$ in \cref{fig:Complete_FRIMU2_JZ_rtm_Trial1_Ascent}. 
	Interestingly at the same time, the locomotion mode recognition predicts SD as well. It seems that the input features at this part of the transition are very similar to the input features of SD, which should be further investigated. 
	
	In contrast to the results, we expected slightly earlier changes in the locomotion mode based on the timing entered by the users with the input unit, as they were instructed to input the transition during the swing phase of the leading limb when entering or leaving the staircase. 
	However, our tests revealed that changes in slope or mode in the majority of cases started after the user-specified transition. Part of this difference is due to the filter used for the visualization and the \SI{300}{\milli \second} time history information used for the input features. In addition we believe that an improved method would be required for a detailed timing analysis with respect to the beginning and end of the human locomotion mode transitions to analyze the changes in the slope and the locomotion mode predictions . Such timings could be determined as performed in \cite{grimmer2020lower}.
	
	Overall the test data results for estimating gait phase during the transitions are very promising. However, based on our analyses we believe that stair ascent in particular would require additional work to improve the detection of the correct slope and locomotion mode during transitions. It must be investigated whether each kind of transition has to be part of the training data as its own locomotion mode or if alternative approaches exist including methods of pre- and post-processing. Furthermore, there is the possibility that the human movement in these transition phases is equal to the steady gait of the incorrectly predicted class and that the biomechanics within this phase are similar. Both theories require further investigation.

	\section{CONCLUSIONS}
	In this work we were able to estimate both gait phase and stair slope as well as predict continuously the locomotion mode for steady level walking and stair ambulation.
	We used artificial neural networks with fully connected layers with input features utilizing time history information. 
	The performance decrease in the real-world test data is lower than that in \cite{Sherratt2021} where an RNN with LSTMs was used.
	
	To obtain test data in real-world environments the FRIMU2 mobile device was developed. With FRIMU2 we could show that the trained ANNs achieved high robustness to different subjects, sensor hardware, sensor fixation, stair height and stair arrangement. 
	
	In addition, we investigated the ability of the network to predict transitions between level walking and stair ambulation for which the network was not trained. It was found that gait phase estimation, stair slop estimation and locomotion mode recognition performed as expected with real-world usable quality for transitions between level walking and stair descent. However, for transitions that include stair ascent additional methods are required to achieve a similar performance as that with stair descent.
	
	One way to improve the performance of our approaches could be individualization of the ANNs to the user to take the user-specific gait into account. With the FRIMU2 setup the necessary data can be collected easily. For subjects with an amputation this could even be done using their prosthesis used in everyday life before using a powered prosthesis with the individualized gait phase and stair slope estimation and locomotion mode recognition as part of the controls.
	
	Based on our results, we believe that the approach can be used to realize a complete high-level lower limb wearable robotic control, in our case specifically for a transtibial powered prosthesis.

	\section*{ACKNOWLEDGMENT}
	
	The authors thank Guoping Zhao for the support in processing the raw data from the laboratory dataset and Philipp Graf for the support in the design of the FRIMU2.
	The authors thank all subjects that took part in the experiments.
	
	\bibliographystyle{IEEEtran}
	\bibliography{IEEEabrv,root}

\end{document}